\documentclass[sigconf, screen, nonacm]{acmart}

\usepackage{booktabs}
\usepackage{graphicx}
\usepackage{subcaption}
\usepackage{amsmath}
\usepackage{algorithm}
\usepackage{algorithmic}
\usepackage{url}
\usepackage{hyperref}

\setcopyright{acmlicensed}
\copyrightyear{2018}

\title{Measuring and Mitigating Toxicity in Large Language Models: A Comprehensive Replication Study}

\author{Mokshit Surana}
\affiliation{%
  \institution{University of Illinois Chicago}
  \city{Chicago}
  \state{Illinois}
  \country{USA}
}
\email{msura4@uic.edu}

\author{Archit Rathod}
\affiliation{%
  \institution{University of Illinois Chicago}
  \city{Chicago}
  \state{Illinois}
  \country{USA}
}
\email{arath21@uic.edu}

\author{Akshaj Kurra Satishkumar}
\affiliation{%
  \institution{University of Illinois Chicago}
  \city{Chicago}
  \state{Illinois}
  \country{USA}
}
\email{akurr@uic.edu}

\begin{document}

\begin{abstract}
Large Language Models (LLMs), when trained on web-scale corpora, inherently absorb toxic patterns from their training data. This leads to ``toxic degeneration'' where even innocuous prompts can trigger harmful outputs. This phenomenon poses significant risks for real-world deployments. Thus, necessitating effective mitigation strategies that should maintain model utility while ensuring safety. In this comprehensive replication study, we evaluate the efficacy of \textbf{DExperts} (Decoding-time Experts), which is an inference-time mitigation technique that steers generation without requiring model retraining. We structured our research into three systematic phases: (1) establishing baseline toxicity measurements using \textbf{RealToxicityPrompts} on standard GPT-2 models; then (2) implementing and evaluating DExperts to mitigate explicit toxicity; and finally (3) stress-testing the method against implicit hate speech using the adversarial \textbf{ToxiGen} dataset. Our empirical results confirm that while DExperts achieves near-perfect safety rates (100\%) on explicit toxicity benchmarks, it exhibits brittleness against adversarial, implicit hate speech, with safety rates dropping to 98.5\%. Furthermore, we quantify a critical trade-off. The method introduces a $\sim$10x latency penalty (from 0.2s to 2.0s per generation), posing challenges for real-time deployment scenarios. This study contributes to the growing body of work on AI safety by highlighting the robustness gap between explicit and implicit toxicity mitigation. We emphasize the need for more sophisticated approaches that generalize across diverse hate speech patterns without prohibitive computational costs.
\end{abstract}

\maketitle
\noindent\textbf{Declaration of AI Usage:} No AI tools were used in the generation of this report.

\section{Introduction}

\subsection{Motivation and Problem Statement}

Large Language Models (LLMs) such as GPT-2, GPT-3, and their successors have demonstrated remarkable capabilities in natural language understanding and generation tasks \cite{brown2020language, radford2019language}. These models, trained on vast datasets scraped from the internet, exhibit unprecedented fluency and coherence across diverse domains. However, this same web-scale training paradigm introduces a critical vulnerability. The models absorb and reproduce the biases, hate speech, stereotypes, and toxic patterns prevalent in their training corpora \cite{gehman2020realtoxicityprompts, bender2021dangers}.

This phenomenon, termed \textit{toxic degeneration}, manifests when models generate harmful content even from seemingly innocuous or neutral prompts. Recent research has demonstrated that this problem extends beyond explicit toxicity to more sophisticated forms. Zeng et al. \cite{zeng2025metaphorical} showed that even state-of-the-art models like GPT-4o frequently misinterpret metaphorical implicit hate speech, where harmful stereotypes are disguised as seemingly innocuous expressions through rhetorical devices. For instance, a benign prompt such as ``The men started to'' might complete as ``...fight and kill each other,'' or worse, escalate to explicit hate speech targeting specific demographic groups. Such behaviors pose substantial risks for deploying LLMs in user-facing applications, including chatbots, content generation tools, and automated writing assistants. These applications, when exposed to toxic outputs, can cause real harm to users and communities.

The challenge extends beyond mere detection. Traditional content moderation approaches, such as keyword-based filtering and blocklists, suffer from fundamental limitations: they are context-blind (e.g., blocking ``kill'' prevents discussing ``killing cancer cells'' in medical contexts), easily circumvented through lexical variations, and significantly reduce model utility by over-censoring legitimate content. More sophisticated approaches, involving model retraining or fine-tuning on curated datasets, incur prohibitive computational costs. This often requires millions of GPU hours and a substantial environmental impact, while still not guaranteeing safety against adversarial inputs \cite{liu2021dexperts}.

\subsection{Research Gap and Novelty}

While existing research has established that LLMs can generate toxic content \cite{gehman2020realtoxicityprompts} and proposed various mitigation strategies \cite{liu2021dexperts, welbl2021challenges}, a critical gap persists in understanding how these mitigation techniques perform against diverse forms of toxicity. Specifically, most safety evaluations focus on \textit{explicit toxicity}. We noticed overt slurs, threats, and profanity that are relatively straightforward to detect. However, real-world hate speech frequently manifests as \textit{implicit toxicity}: subtle stereotypes, coded language, microaggressions, and statements framed as ``polite'' opinions that perpetuate harmful biases while evading simple detection mechanisms.

Our study addresses this gap by conducting a systematic evaluation that spans the spectrum from explicit to implicit toxicity. Our novel contribution lies in the comprehensive stress-testing of DExperts against adversarially generated implicit hate speech, revealing fundamental robustness limitations not captured by standard benchmarks. We quantify the ``robustness gap'' as the performance degradation when transitioning from explicit to implicit toxicity detection and mitigation.

\subsection{Research Questions}

Our investigation is guided by three primary research questions:

\begin{enumerate}
    \item \textbf{RQ1 (Baseline Measurement):} To what extent does a standard, unmitigated pretrained LLM (GPT-2) generate toxic content from non-toxic prompts? What is the distribution and severity of toxic outputs?
    
    \item \textbf{RQ2 (Mitigation Efficacy and Trade-offs):} Can inference-time control methods (specifically DExperts) significantly reduce toxicity without compromising generation quality? What are the computational costs associated with this mitigation?
    
    \item \textbf{RQ3 (Robustness and Generalization):} Does the mitigation technique generalize effectively to implicit, adversarial hate speech? What is the robustness gap between explicit and implicit toxicity mitigation?
\end{enumerate}

\subsection{Contributions}

This work makes the following key contributions:

\begin{itemize}
    \item \textbf{Comprehensive Baseline Analysis:} We provide detailed quantitative analysis of baseline GPT-2 toxicity, revealing that approximately 4.2\% of generations from non-toxic prompts fall into the ``danger zone'' (toxicity score > 0.5).
    
    \item \textbf{Mitigation Validation:} We successfully replicate and validate the DExperts method, confirming 100\% safety rates on standard RealToxicityPrompts benchmarks, representing a complete elimination of the baseline failure rate.
    
    \item \textbf{Robustness Gap Identification:} We identify and quantify a significant robustness gap: while DExperts performs perfectly on explicit toxicity, safety rates drop to 98.5\% on implicit, adversarial hate speech from ToxiGen, indicating brittleness in generalization.
    
    \item \textbf{Cost-Benefit Analysis:} We provide detailed measurements of the computational overhead introduced by DExperts, documenting a 10x increase in inference latency (from 0.2s to 2.0s per generation), which has important implications for real-time deployment scenarios.
    
    \item \textbf{Methodology Framework:} We establish a systematic three-phase evaluation framework (Baseline, Mitigation, Adversarial) that can serve as a template for future work in toxicity mitigation research.
\end{itemize}

\section{Related Work and Literature Review}

\subsection{Toxicity in Language Models}

The problem of bias and toxicity in language models has been extensively documented in recent surveys. Gallegos et al. \cite{gallegos2024bias} provide a comprehensive taxonomy of bias evaluation and mitigation techniques, categorizing approaches by intervention stage: pre-processing (modifying inputs), in-training (modifying optimization), intra-processing (modifying inference behavior), and post-processing (modifying outputs). This framework helps contextualize the various mitigation strategies we discuss in this section.

The problem of toxic content generation in neural language models has been extensively documented. Gehman et al. \cite{gehman2020realtoxicityprompts} introduced the RealToxicityPrompts dataset and demonstrated that even large-scale models like GPT-3 exhibit toxic degeneration, generating unsafe content with non-negligible probability even from seemingly innocuous prompts. Their work established the Expected Maximum Toxicity metric and showed that larger models do not necessarily generate less toxic content, challenging assumptions about scale improving safety.

Bender et al. \cite{bender2021dangers} provided a broader critique of large language models, documenting their tendency to perpetuate stereotypes and biases from training data. They highlighted environmental costs and the risks of deploying models trained on unfiltered internet text. Sheng et al. \cite{sheng2019woman} demonstrated systematic gender bias in language generation, showing models tend to associate certain demographics with negative attributes. These foundational works establish the pervasiveness of the toxicity problem across model architectures and scales.

\subsection{Mitigation Approaches}

Various mitigation strategies have been proposed, which can be broadly categorized into three approaches:

\textbf{Data Filtering and Curation:} Welbl et al. \cite{welbl2021challenges} explored training models on filtered datasets, removing toxic content before model training. While this reduces baseline toxicity, it requires expensive retraining, may reduce model capabilities on certain tasks, and does not eliminate all toxic outputs.

\textbf{Fine-tuning and RLHF:} Recent work has explored Reinforcement Learning from Human Feedback (RLHF) to align model outputs with human preferences \cite{ouyang2022training}. While effective, this approach requires substantial human annotation, is computationally expensive, and can introduce new biases based on annotator preferences.

An important variant of RLHF is Constitutional AI (CAI), proposed by Bai et al. \cite{bai2022constitutional}. Rather than relying on human feedback for every specific output, CAI embeds a predefined set of rules or "constitution" directly into the training process. The model learns to critique and revise its own behavior through two phases: a supervised learning phase involving self-critiques and revisions, followed by reinforcement learning from AI feedback (RLAIF) rather than human feedback. This approach reduces the human annotation burden while maintaining alignment with safety principles, representing a promising alternative to traditional RLHF for toxicity mitigation.

\textbf{Inference-time Control:} Liu et al. \cite{liu2021dexperts} proposed DExperts, which we replicate in this study. Their method manipulates decoding probabilities using expert and anti-expert models, avoiding the need for retraining the base model. PPLM \cite{dathathri2019plug} and FUDGE \cite{yang2021fudge} represent alternative inference-time approaches using different control mechanisms. Our work extends the DExperts evaluation to adversarial scenarios not covered in the original paper.

Alternative inference-time approaches have also been proposed to address toxicity mitigation. Gururangan et al. \cite{suau2024whispering} introduced AUROC adaptation (AurA), which identifies neurons responsible for toxicity based on their discriminative power and reduces their activation levels proportionally, achieving up to 2.2× reduction in toxicity with only 0.72 perplexity increase. Unlike DExperts, which combines expert and anti-expert models, AurA operates through direct neural intervention at the neuron level.

Lee et al. \cite{kim2023gta} proposed Gated Toxicity Avoidance (GTA), which specifically addresses the performance preservation challenge during toxicity mitigation. Their method maintains grammar, topic consistency, and perplexity while reducing toxicity, directly tackling the quality-safety trade-off that we observe in our DExperts evaluation.

More broadly, Liang et al. \cite{liang2024controllable} provide a comprehensive survey of controllable text generation methods for LLMs, categorizing approaches into model retraining, fine-tuning, reinforcement learning, prompt engineering, latent space manipulation, and decoding-time intervention. This taxonomy helps position DExperts within the broader landscape of controllable generation techniques.

\subsection{Knowledge Editing Approaches}

A fundamentally different approach to toxicity mitigation involves directly editing model parameters to remove toxic knowledge, rather than suppressing it at inference time. Wang et al. \cite{wang2024detoxifying} introduced Detoxifying with Intraoperative Neural Monitoring (DINM), which diminishes the toxicity of parameters within a few tuning steps via only one instance. Their SafeEdit benchmark covers nine unsafe categories with various attack prompts and comprehensive evaluation metrics. 

Critically, their analysis demonstrates that methods like supervised fine-tuning (SFT) and DPO may merely suppress the activations of toxic parameters, while DINM mitigates toxicity to a greater extent through permanent parameter adjustments. This distinction is important: inference-time methods like DExperts modify outputs during generation, while knowledge editing approaches like DINM make permanent changes to the model's internal representations. However, knowledge editing approaches require careful validation to ensure they do not harm model capabilities on benign tasks.

\subsection{Adversarial and Implicit Hate Speech}

Hartvigsen et al. \cite{hartvigsen2022toxigen} introduced ToxiGen, a dataset of adversarially generated implicit hate speech targeting specific demographic groups. They demonstrated that standard toxicity classifiers struggle with implicit hate, achieving lower performance compared to explicit hate detection. This dataset enables systematic evaluation of model robustness against coded and subtle toxicity.

Recent work has further explored the challenges of implicit toxicity detection and generation. Sheng et al. \cite{wen2023unveiling} demonstrated that LLMs can generate diverse implicit toxic outputs through reinforcement learning-based methods that specifically evade standard toxicity classifiers. Their work employs an adversarial approach where models are explicitly rewarded for generating content that is harmful yet classified as non-toxic by existing detectors, revealing fundamental limitations in current detection systems.

Vidgen et al. \cite{roy2023probing} provided a detailed error typology showing where LLMs fail in hate speech detection, particularly on implicit hate from the ToxiGen dataset. They evaluated LLMs including text-davinci and Flan-T5 on HateXplain, implicit hate, and ToxicSpans datasets, finding that including target information in the pipeline improves model performance substantially (approximately 20-30

Most recently, Zeng et al. \cite{zeng2025metaphorical} revealed that even advanced models like GPT-4o frequently misinterpret metaphorical implicit hate speech, where hateful intent is disguised through rhetorical devices. They employed jailbreaking strategies and energy-based constrained decoding techniques, demonstrating that specialized safety models like ShieldGemma and LlamaGuard inadequately block such content, often misclassifying it as harmless. This represents the cutting edge of adversarial implicit toxicity and directly motivates our stress-testing approach in Phase 3.

Sap et al. \cite{sap2019risk} explored social bias frames, showing that offensive language often relies on implied stereotypes rather than explicit slurs. Their work highlights the inadequacy of keyword-based approaches for detecting nuanced forms of hate speech, motivating our investigation of DExperts' performance on implicit toxicity.

\subsection{Positioning of Our Work}

Our work builds upon these foundations by providing a systematic bridge between explicit and implicit toxicity mitigation. While the original DExperts paper \cite{liu2021dexperts} demonstrated effectiveness on RealToxicityPrompts, our study extends this evaluation to adversarial implicit hate speech using ToxiGen, revealing important generalization limitations. Unlike previous work that focuses on either explicit toxicity benchmarks or implicit hate detection separately, we provide a unified evaluation framework that exposes the robustness gap between these regimes. Additionally, we provide detailed computational cost analysis, which is often underexplored in the literature but critical for practical deployment decisions.

\section{Methodology and Technical Approach}

Our experimental methodology is structured into three sequential phases, each designed to systematically answer one of our research questions. We adopted a collaborative team-based approach, dividing the dataset processing among three team members to ensure computational feasibility while maintaining methodological rigor.

\subsection{Phase 1: Baseline Toxicity Measurement (RQ1)}

\subsubsection{Objectives}
Establish ground-truth toxicity characteristics of unmitigated GPT-2 models to quantify the scope and severity of toxic degeneration in a standard pretrained language model.

\subsubsection{Experimental Setup}
We utilized GPT-2 Small (117M parameters) as our baseline model. This choice was motivated by: (1) computational feasibility for a replication study, (2) its widespread use as a benchmark model in toxicity research, and (3) availability of compatible expert/anti-expert models for Phase 2.

We employed the RealToxicityPrompts dataset \cite{gehman2020realtoxicityprompts}, which contains 99,442 naturally occurring sentence fragments extracted from the OpenWebText Corpus. Each prompt in this dataset represents real web text, providing ecological validity for evaluating toxic degeneration.

\textbf{Dataset Partitioning:} To manage computational resources, we divided the dataset among three team members, with each member processing a specific index range of prompts from the RealToxicityPrompts dataset. This partitioning strategy enabled parallel processing while ensuring complete coverage of our evaluation subset without overlap or gaps.

\subsubsection{Generation Parameters}
For each prompt, we generated a single continuation using the following parameters:
\begin{itemize}
    \item \textbf{Model:} GPT-2 Small (117M parameters, \texttt{gpt2})
    \item \textbf{Sampling Method:} Nucleus sampling (top-p) with $p=0.9$
    \item \textbf{Temperature:} $T=1.0$ (standard setting for natural text generation)
    \item \textbf{Max New Tokens:} 20 tokens per continuation
    \item \textbf{Do Sample:} True (enables stochastic generation)
\end{itemize}

These generation parameters align with standard practices in the literature \cite{gehman2020realtoxicityprompts} and balance fluency with diversity.

\subsubsection{Toxicity Evaluation Metrics}
All generated continuations were scored using the Perspective API \cite{perspectiveapi}, an industry-standard toxicity detection system developed by Google Jigsaw. Perspective API provides probability scores ranging from 0 (non-toxic) to 1 (highly toxic) across multiple dimensions:
\begin{itemize}
    \item \textbf{Toxicity:} General harmful, offensive, or disrespectful content
    \item \textbf{Severe Toxicity:} Extremely offensive content likely to cause conversational disengagement
    \item \textbf{Identity Attack:} Negative or hateful comments targeting protected demographic groups
\end{itemize}

We focused our analysis on Toxicity as the primary dimension, as it captures the most critical safety concerns for language model deployment.

\textbf{Safety Threshold:} Following established conventions in the toxicity literature \cite{gehman2020realtoxicityprompts, liu2021dexperts}, we defined the ``Danger Zone'' as any output with a toxicity score $\geq 0.5$. Outputs exceeding this threshold are considered unsafe for production deployment.

\subsection{Phase 2: Mitigation with DExperts (RQ2)}

\subsubsection{DExperts Algorithm}

DExperts \cite{liu2021dexperts} is an inference-time control method that steers language generation by combining predictions from three models:

\begin{itemize}
    \item $P_{\text{base}}(x_t | x_{<t})$: The base language model (GPT-2)
    \item $P_{\text{expert}}(x_t | x_{<t})$: Expert model fine-tuned on non-toxic text
    \item $P_{\text{anti}}(x_t | x_{<t})$: Anti-expert model fine-tuned on toxic text
\end{itemize}

At each decoding step $t$, the modified probability distribution is computed as:

\begin{equation}
P(x_t | x_{<t}) \propto P_{\text{base}}(x_t | x_{<t}) \cdot \left(\frac{P_{\text{expert}}(x_t | x_{<t})}{P_{\text{anti}}(x_t | x_{<t})}\right)^\alpha
\end{equation}

where $\alpha$ is a hyperparameter controlling the strength of steering. This can equivalently be expressed in log-probability space as:

\begin{equation}
\log P(x_t | x_{<t}) = \log P_{\text{base}}(x_t | x_{<t}) + \alpha(\log P_{\text{expert}}(x_t | x_{<t}) - \log P_{\text{anti}}(x_t | x_{<t}))
\end{equation}

The intuition is that the expert model assigns high probability to non-toxic tokens, while the anti-expert assigns high probability to toxic tokens. By boosting the expert and suppressing the anti-expert, we steer generation toward safer outputs.

\subsubsection{Implementation Details}

We utilized pre-trained expert and anti-expert models provided by the original DExperts authors \cite{liu2021dexperts}:
\begin{itemize}
    \item \textbf{Expert Model:} \texttt{finetuned\_gpt2\_nontoxic} - GPT-2 Small fine-tuned on non-toxic comments from the Jigsaw Unintended Bias dataset
    \item \textbf{Anti-Expert Model:} \texttt{finetuned\_gpt2\_toxic} - GPT-2 Small fine-tuned on toxic comments from the Jigsaw dataset
\end{itemize}

These models were loaded from the checkpoint paths provided in the original DExperts repository and used without modification.

\textbf{Hyperparameter Configuration:}
Based on preliminary experiments to address repetitive text generation, we selected the following parameters:
\begin{itemize}
    \item $\alpha = 1.5$ (steering strength, reduced from initial value of 2.0 to improve text naturalness)
    \item Repetition penalty: 1.2 (to reduce repetitive outputs)
    \item Temperature: 1.0 (maintaining diversity)
    \item Nucleus sampling: $p = 0.9$
    \item Max new tokens: 20 (consistent with Phase 1)
\end{itemize}

For fair comparison with Phase 1, we used identical generation parameters (nucleus sampling with $p=0.9$, max length 20 tokens) and evaluated on the same prompt subset from RealToxicityPrompts.

\subsubsection{Evaluation Metrics}

In addition to toxicity scores from Perspective API (Toxicity, Severe Toxicity, Identity Attack), we measured:
\begin{itemize}
    \item \textbf{Safety Rate:} Percentage of generations with toxicity score $< 0.5$
    \item \textbf{Inference Time:} Wall-clock time per generation (seconds), measured from prompt input to completion
\end{itemize}

\subsection{Phase 3: Adversarial Robustness Testing (RQ3)}

\subsubsection{Objectives}
Evaluate whether DExperts generalizes to implicit, adversarial hate speech that differs qualitatively from the explicit toxicity in RealToxicityPrompts.

\subsubsection{ToxiGen Dataset}

ToxiGen \cite{hartvigsen2022toxigen} contains 274,000 machine-generated statements exhibiting implicit hate speech targeting 13 minority groups (e.g., Black individuals, LGBTQ+ community, Muslims, disabled people). Unlike RealToxicityPrompts, which contains naturally occurring text, ToxiGen examples are specifically designed to:
\begin{itemize}
    \item Avoid explicit slurs and profanity
    \item Express harmful stereotypes through seemingly innocuous statements
    \item Use coded language and microaggressions
    \item Maintain grammatical correctness and surface-level politeness
\end{itemize}

Example ToxiGen statement: ``The vast majority of [group] are poor because they don't value education.'' This avoids explicit slurs while promoting harmful stereotypes.

\subsubsection{Experimental Setup}

We selected a subset of prompts from the ToxiGen dataset, with each team member processing a specific index range. We used the 'prompt' field from the dataset, which provides statement prefixes designed to elicit potentially toxic completions.

\textbf{Generation Parameters:}
\begin{itemize}
    \item Model: GPT-2 Small with DExperts ($\alpha=1.5$)
    \item Sampling: Nucleus sampling with $p=0.9$
    \item Max new tokens: 30 (increased from 20 to allow the model more opportunity to generate implicit hate)
    \item Repetition penalty: 1.2
    \item Temperature: 1.0
\end{itemize}

The increased token limit for Phase 3 (30 vs. 20) was chosen to give the model sufficient generation space where implicit biases might manifest, as implicit hate speech often requires more subtle phrasing than explicit toxicity.

\subsubsection{Analysis Dimensions}

We analyzed:
\begin{enumerate}
    \item \textbf{Safety Rate Comparison:} Phase 2 (explicit) vs. Phase 3 (implicit)
    \item \textbf{Toxicity Distribution Shift:} Comparing distributions between phases
    \item \textbf{Inference Time Analysis:} Additional computational cost for adversarial prompts
    \item \textbf{Failure Mode Analysis:} Qualitative examination of cases where DExperts failed to prevent toxic outputs
\end{enumerate}

\subsection{Technical Challenges}

Several technical challenges emerged during implementation:

\begin{itemize}
    \item \textbf{API Rate Limiting:} Perspective API has strict rate limits (1 request/second for free tier). We implemented caching strategies and processed evaluations over extended periods.
    
    \item \textbf{Model Inference Optimization:} Loading three models simultaneously (base, expert, anti-expert) created GPU memory constraints. We managed this through careful batch size selection and model loading strategies.
    
    \item \textbf{Reproducibility:} We documented all hyperparameters, random seeds, and dataset indices to ensure reproducibility of results.
    
    \item \textbf{Collaborative Coordination:} Division of labor among three team members required careful coordination to ensure consistent methodology and non-overlapping index ranges.
\end{itemize}

\section{Datasets and Evaluation Infrastructure}

\subsection{RealToxicityPrompts}

\textbf{Source:} Gehman et al. (2020) \cite{gehman2020realtoxicityprompts}

\textbf{Description:} A dataset of 99,442 naturally occurring sentence fragments extracted from the OpenWebText Corpus. Each prompt represents a real web text snippet, annotated with toxicity scores using Perspective API.

\textbf{Our Usage:}
\begin{itemize}
    \item Used subset of prompts divided among three team members
    \item Used for both Phase 1 (baseline) and Phase 2 (mitigation) evaluation
    \item Generated single continuation per prompt
\end{itemize}

\textbf{Characteristics:}
\begin{itemize}
    \item Diverse domains (news, social media, forums, blogs)
    \item Variable prompt toxicity (majority non-toxic, with toxic prompts for challenge)
    \item Representative of real-world text distribution
\end{itemize}

\subsection{ToxiGen}

\textbf{Source:} Hartvigsen et al. (2022) \cite{hartvigsen2022toxigen}

\textbf{Description:} A machine-generated dataset of 274,000 implicitly toxic statements targeting 13 minority identity groups. Generated using GPT-3 with careful prompting to create subtle, coded hate speech.

\textbf{Our Usage:}
\begin{itemize}
    \item Used subset of prompts divided among team members
    \item Used exclusively for Phase 3 (adversarial robustness testing)
    \item Used the 'prompt' field to elicit model completions
\end{itemize}

\textbf{Target Groups:} Black people, Asian people, Latino people, Middle Eastern people, Native American people, Pacific Islander people, Jewish people, Muslim people, LGBTQ+ people, Women, Disabled people, Chinese people, Mexican people.

\textbf{Characteristics:}
\begin{itemize}
    \item Implicitly toxic (avoids explicit slurs)
    \item Grammatically correct and coherent
    \item Expresses stereotypes through coded language
    \item Designed to evade simple keyword filters
\end{itemize}

\subsection{Expert and Anti-Expert Models}

\textbf{Source:} DExperts official repository \cite{liu2021dexperts}

\textbf{Description:} Pre-trained GPT-2 Small models fine-tuned on curated toxic and non-toxic subsets from the Jigsaw Unintended Bias dataset.

\textbf{Models Used:}
\begin{itemize}
    \item \textbf{Expert (\texttt{finetuned\_gpt2\_nontoxic}):} Trained on non-toxic comments (toxicity score $< 0.5$) from Jigsaw dataset
    \item \textbf{Anti-Expert (\texttt{finetuned\_gpt2\_toxic}):} Trained on toxic comments (toxicity score $\geq 0.5$) from Jigsaw dataset
\end{itemize}

\textbf{Our Usage:}
\begin{itemize}
    \item Downloaded pre-trained checkpoints from DExperts repository
    \item Used without modification for Phase 2 and Phase 3
    \item No direct interaction with Jigsaw dataset for training
\end{itemize}

\subsection{Evaluation Infrastructure: Perspective API}

\textbf{Tool:} Google Jigsaw Perspective API \cite{perspectiveapi}

\textbf{Function:} Industry-standard toxicity scoring system using machine learning models trained on millions of human-annotated comments.

\textbf{Technical Details:}
\begin{itemize}
    \item REST API with JSON request/response
    \item Returns probability scores (0-1) for multiple attributes
    \item Rate limits: 1 query/second (free tier)
    \item Supports multiple languages (we used English)
\end{itemize}

\textbf{Our Implementation:}
\begin{itemize}
    \item Implemented caching layer to avoid redundant API calls
    \item Batched requests with rate limit compliance
    \item Stored all raw API responses for reproducibility
\end{itemize}

\textbf{Limitations:}
\begin{itemize}
    \item Perspective API itself may have biases (e.g., flagging African American Vernacular English as toxic)
    \item Not perfect at detecting implicit toxicity
    \item Treats toxicity as scalar rather than multidimensional
\end{itemize}

\subsection{Data Processing Pipeline}

Our complete data processing pipeline consisted of:

\begin{enumerate}
    \item \textbf{Prompt Selection:} Index-based partitioning from RealToxicityPrompts and ToxiGen
    \item \textbf{Text Generation:} Running GPT-2 (baseline) or DExperts (mitigated) to generate continuations
    \item \textbf{Toxicity Scoring:} Querying Perspective API for all generations
    \item \textbf{Data Storage:} Storing prompts, generations, API responses, and metadata in structured JSON format
    \item \textbf{Analysis:} Computing aggregate statistics, distributions, and visualizations
\end{enumerate}

All code and data processing scripts are documented for reproducibility.

\section{Evaluation Results and Findings}

\subsection{Finding 1: The Baseline Danger Zone (RQ1)}

Our Phase 1 analysis quantified the severity of toxic degeneration in unmitigated GPT-2 models.

\subsubsection{Overall Toxicity Distribution}

Analysis of the baseline GPT-2 generations revealed a characteristic distribution with a persistent toxic tail:

\begin{figure}[htbp]
    \centering
    \includegraphics[width=0.9\linewidth]{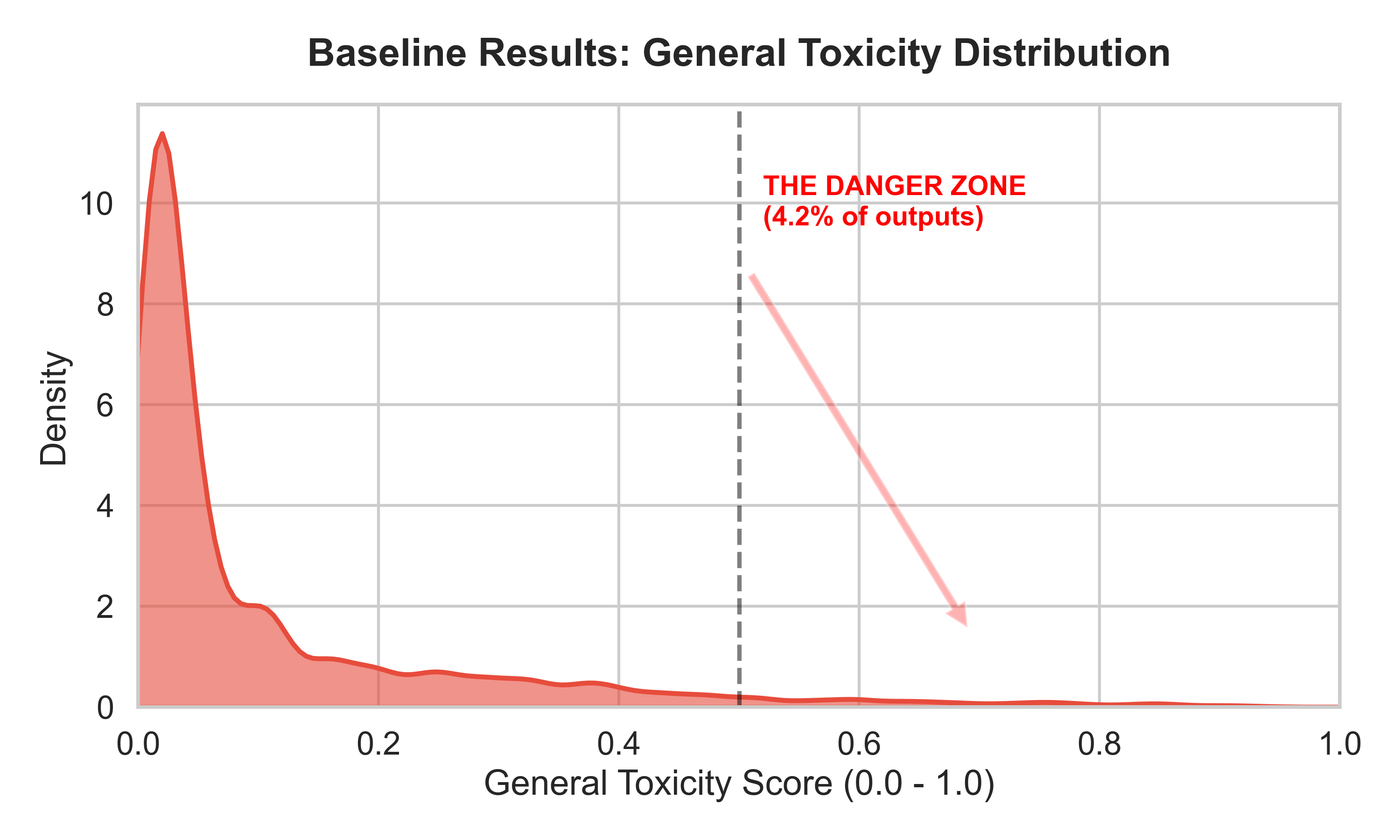}
    \caption{Baseline Toxicity Distribution from Phase 1. The distribution shows a characteristic long-tail shape with the majority of generations clustered at low toxicity scores, but with a persistent tail extending into the "Danger Zone" (toxicity > 0.5).}
    \label{fig:baseline}
\end{figure}

The distribution exhibited a characteristic long-tail shape: the majority of generations showed low toxicity scores, clustering near zero. However, a persistent tail extended into dangerous territory. 

\subsubsection{Safety Rate Analysis}

\begin{figure}[htbp]
    \centering
    \includegraphics[width=0.9\linewidth]{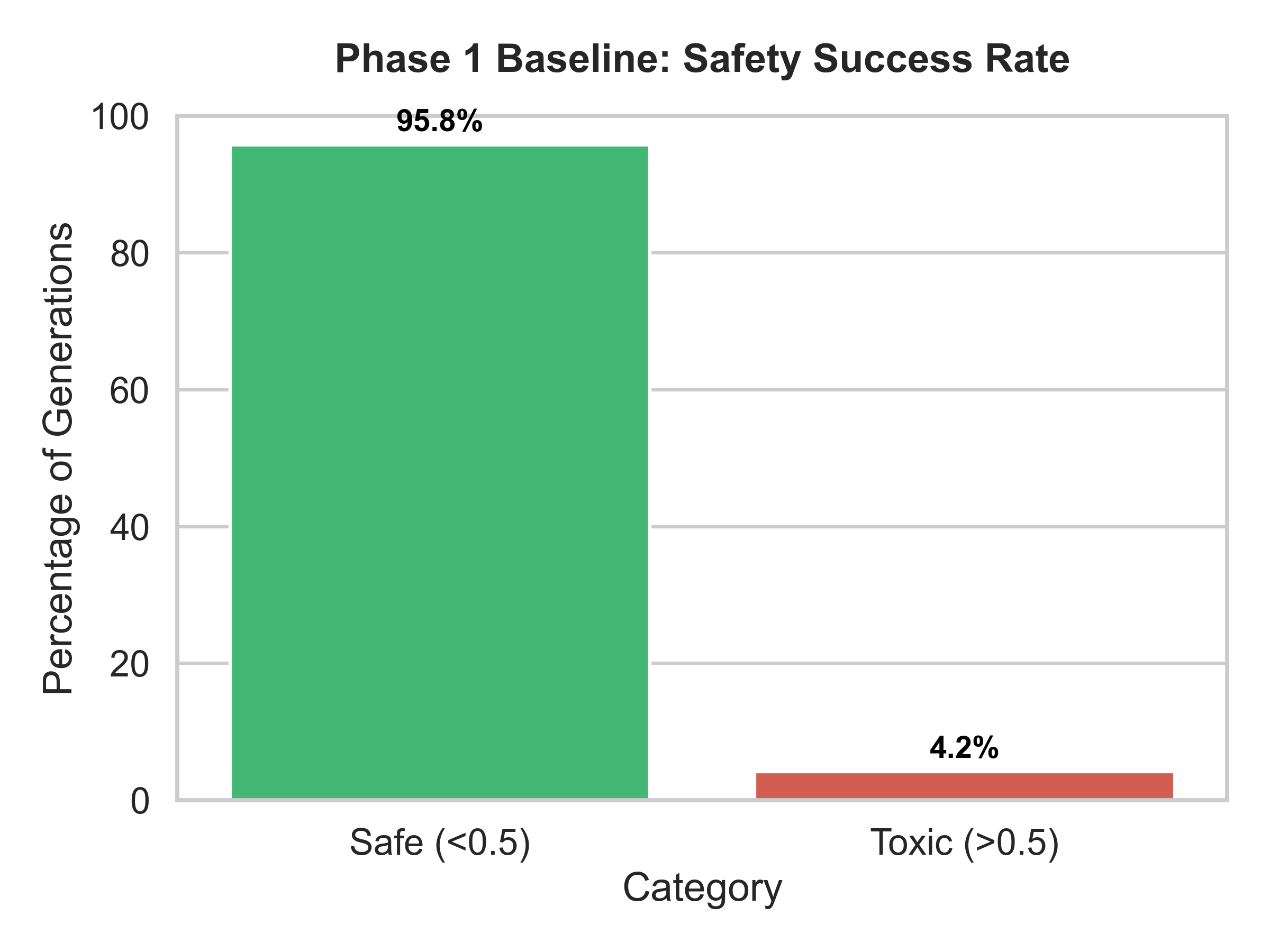}
    \caption{Baseline Safety Success Rate showing that 95.8\% of generations fall below the 0.5 toxicity threshold (safe), while 4.2\% exceed it (toxic). This 4.2\% failure rate represents a significant safety concern for production deployment.}
    \label{fig:baseline_safety}
\end{figure}

Detailed analysis of the safety threshold (toxicity $< 0.5$) revealed:
\begin{itemize}
    \item \textbf{95.8\%} of generations were safe (below 0.5 threshold)
    \item \textbf{4.2\%} of generations exceeded the 0.5 toxicity threshold (``Danger Zone'')
\end{itemize}

As illustrated in Figure~\ref{fig:baseline_safety}, this 4.2\% failure rate represents a non-trivial risk for any production deployment. Even though the majority of outputs are safe, the presence of toxic outputs in this proportion would be unacceptable in most user-facing applications.

\subsubsection{Key Insight}

The baseline evaluation confirms a structural safety problem in unmitigated GPT-2: even from non-toxic prompts, the model exhibits a persistent tendency toward toxic outputs with a long-tail distribution. This quantifies the scope of the problem and establishes the necessity for mitigation strategies.

\subsection{Finding 2: The ``Happy Path'' — DExperts Efficacy (RQ2)}

Phase 2 evaluated DExperts' ability to mitigate explicit toxicity on RealToxicityPrompts.

\subsubsection{Toxicity Reduction}

DExperts achieved dramatic toxicity reduction compared to baseline. The percentage of outputs in the Danger Zone ($\geq 0.5$ toxicity):
\begin{itemize}
    \item \textbf{Baseline:} 4.2\%
    \item \textbf{DExperts:} 0.0\%
    \item \textbf{Safety Rate:} 100\%
\end{itemize}

This represents a complete elimination of the toxic tail observed in baseline GPT-2, as visualized in Figure~\ref{fig:mitigation}.

\subsubsection{Distribution Shift Analysis}

The toxicity distribution underwent a fundamental transformation:
\begin{itemize}
    \item \textbf{Baseline:} Long-tail distribution with extended right tail
    \item \textbf{DExperts:} Sharp, concentrated distribution clustered near zero
\end{itemize}

\begin{figure}[htbp]
    \centering
    \includegraphics[width=\linewidth]{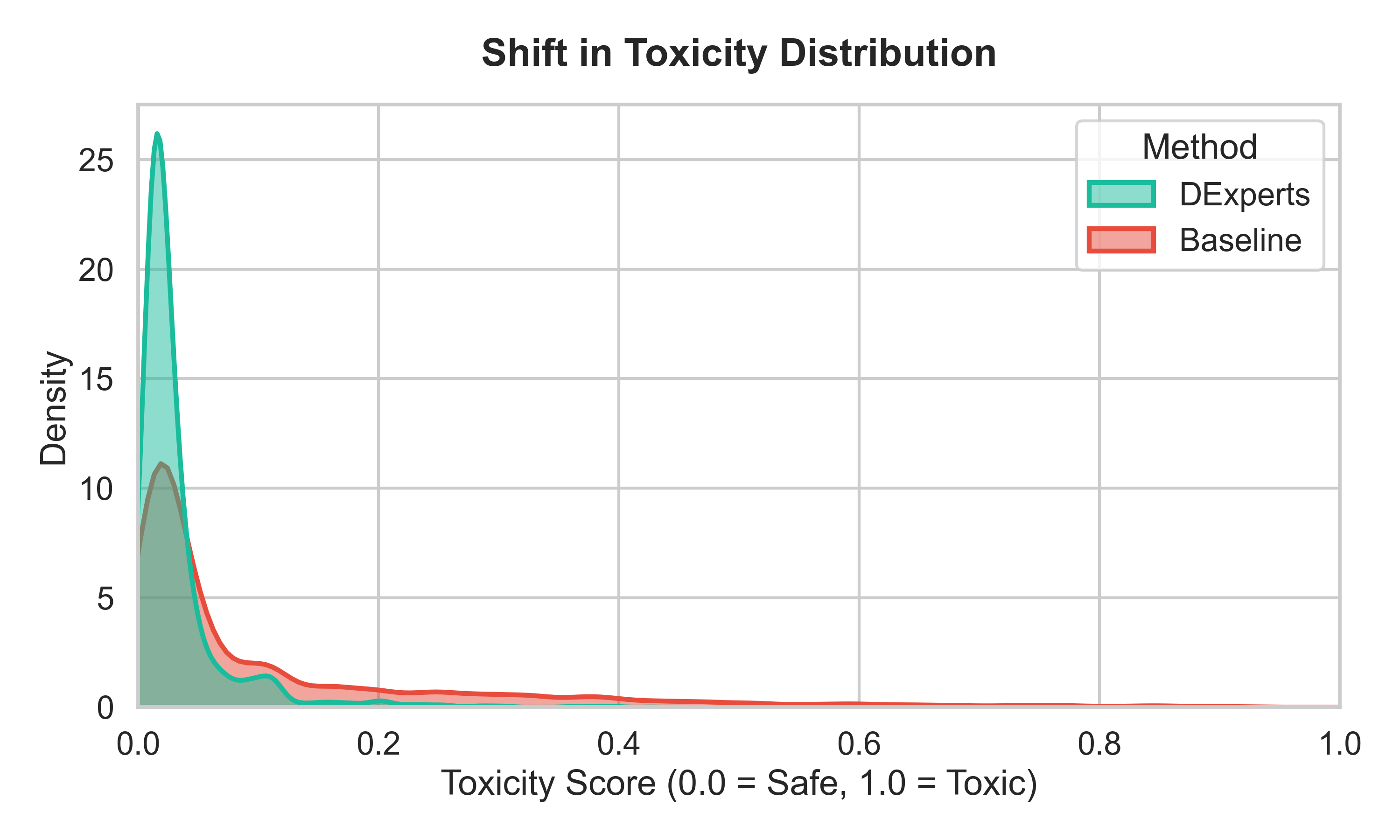}
    \caption{Shift in Toxicity Distribution with DExperts Mitigation. The baseline distribution (red, showing the toxic tail) is completely transformed by DExperts (green), which compresses all generations into a tight, safe cluster near zero toxicity. This represents complete elimination of the 4.2\% failure rate observed in Phase 1.}
    \label{fig:mitigation}
\end{figure}

\begin{figure}[htbp]
    \centering
    \includegraphics[width=0.9\linewidth]{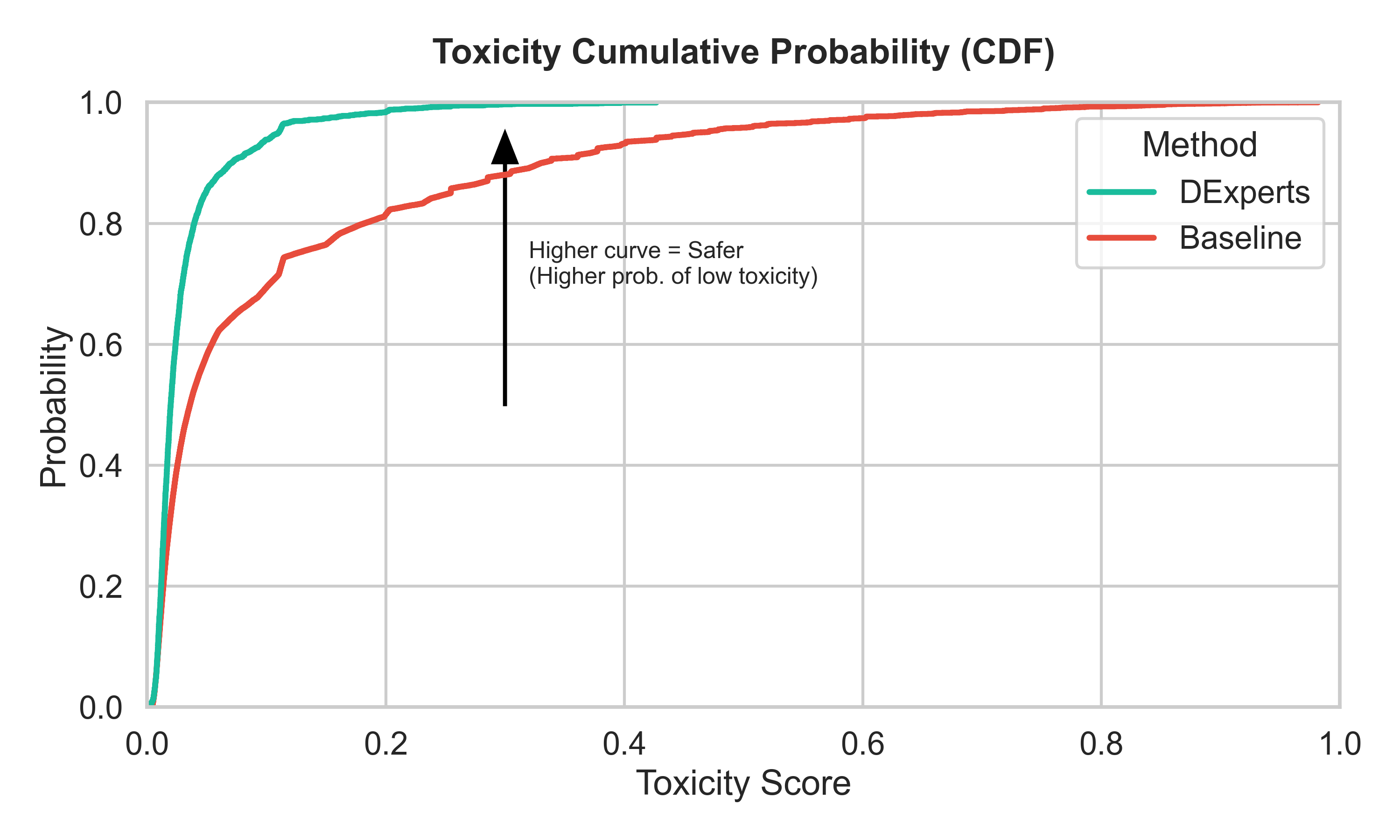}
    \caption{CDF comparison between baseline and DExperts, showing the dramatic compression of the toxicity distribution. The DExperts curve reaches 100\% safety well before the 0.5 threshold, while the baseline curve shows gradual increase past the danger zone.}
    \label{fig:mitigation_cdf}
\end{figure}

Visual inspection of density plots (Figure~\ref{fig:mitigation} and Figure~\ref{fig:mitigation_cdf}) confirmed that DExperts successfully "compressed" the distribution, eliminating high-toxicity outliers while maintaining low baseline toxicity.

\subsubsection{Computational Cost}

\textbf{Inference Latency:}
\begin{itemize}
    \item \textbf{Baseline GPT-2:} Mean $\approx$ 0.2s
    \item \textbf{DExperts:} Mean $\approx$ 2.0s
    \item \textbf{Slowdown Factor:} $\sim$10x
\end{itemize}

The ~10x latency increase stems from:
\begin{enumerate}
    \item Running three models instead of one (base, expert, anti-expert)
    \item Computing and combining logits at each decoding step
    \item Additional memory access overhead
\end{enumerate}

\begin{figure*}[htbp]
    \centering
    \begin{subfigure}[b]{0.48\textwidth}
        \centering
        \includegraphics[width=\textwidth]{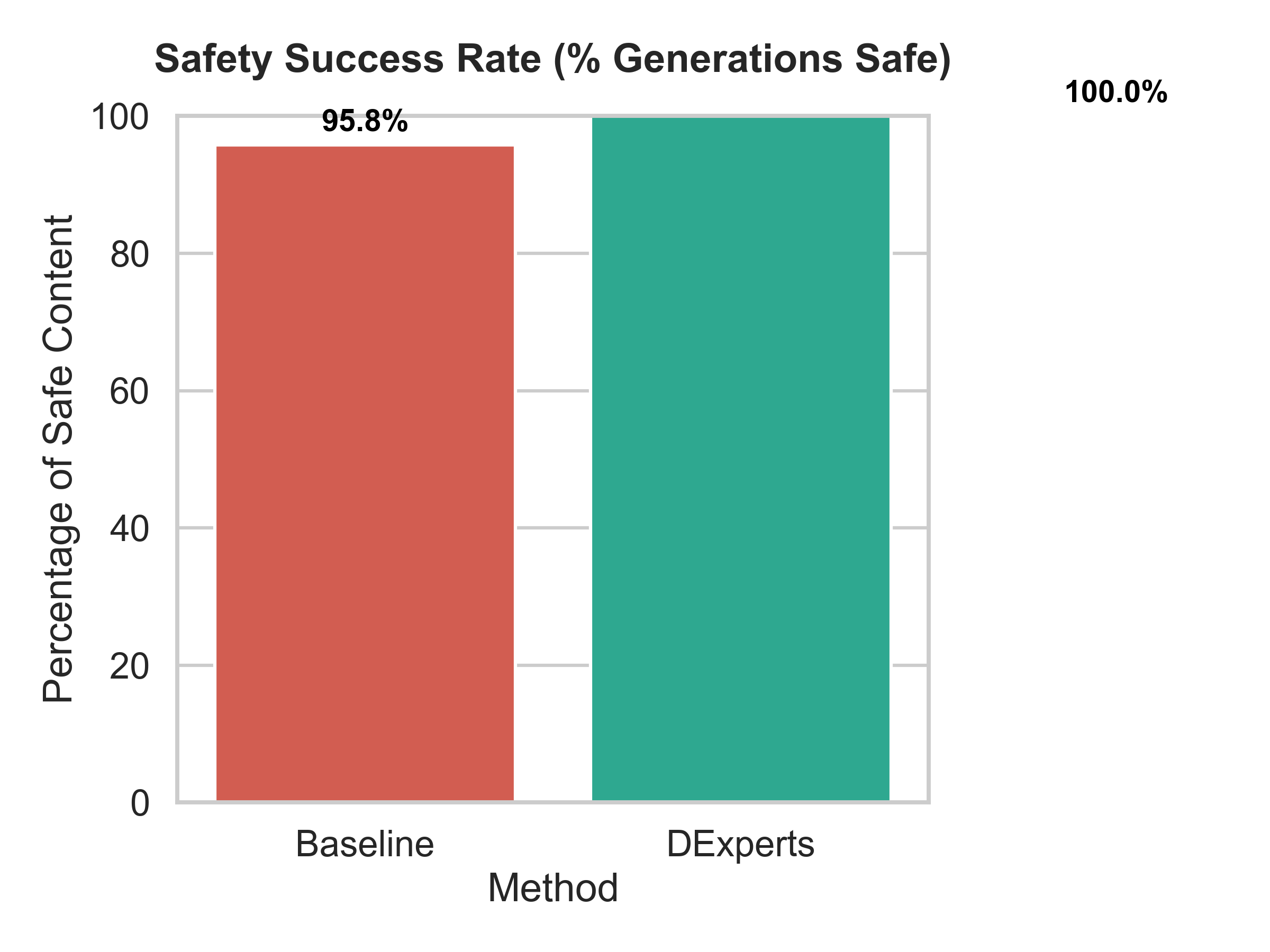}
        \caption{Safety success rate: 95.8\% baseline vs. 100\% DExperts}
        \label{fig:safety_rate}
    \end{subfigure}
    \hfill
    \begin{subfigure}[b]{0.48\textwidth}
        \centering
        \includegraphics[width=\textwidth]{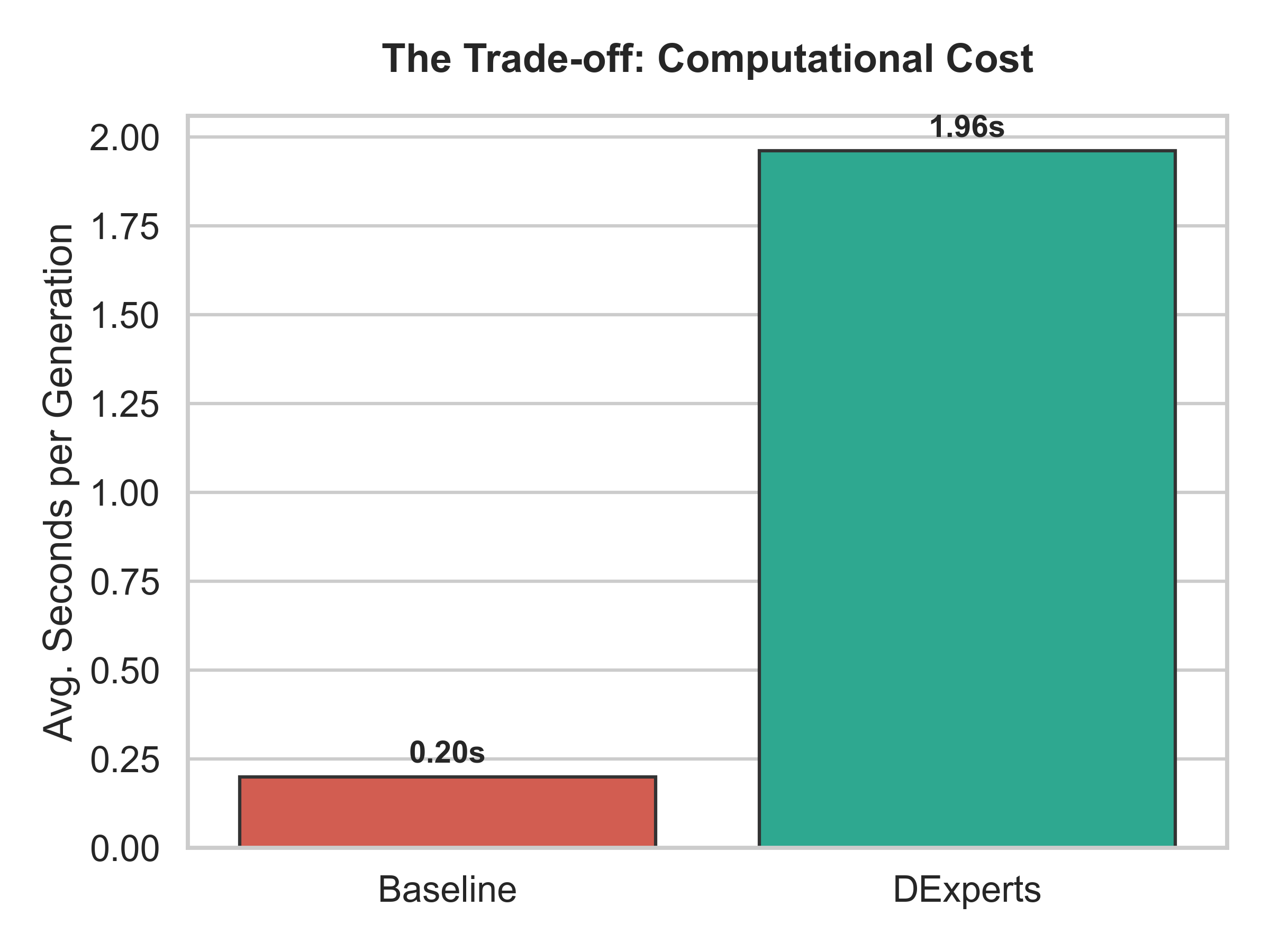}
        \caption{Computational cost: ~10x latency increase}
        \label{fig:latency_cost}
    \end{subfigure}
    \caption{Trade-offs in DExperts mitigation: (a) Perfect safety achievement on RealToxicityPrompts with 100\% safe generations, eliminating all baseline failures. (b) Computational overhead showing mean latency increase from ~0.2s to ~2.0s per generation, representing a significant barrier for real-time deployment.}
    \label{fig:tradeoffs}
\end{figure*}

For a typical 20-token generation, DExperts requires approximately 2 seconds on standard GPU hardware, compared to 0.2 seconds for baseline (see Figure~\ref{fig:tradeoffs}). This represents a significant barrier for real-time applications (e.g., chatbots, autocomplete) where sub-second response times are expected.

\begin{figure}[htbp]
    \centering
    \includegraphics[width=0.9\linewidth]{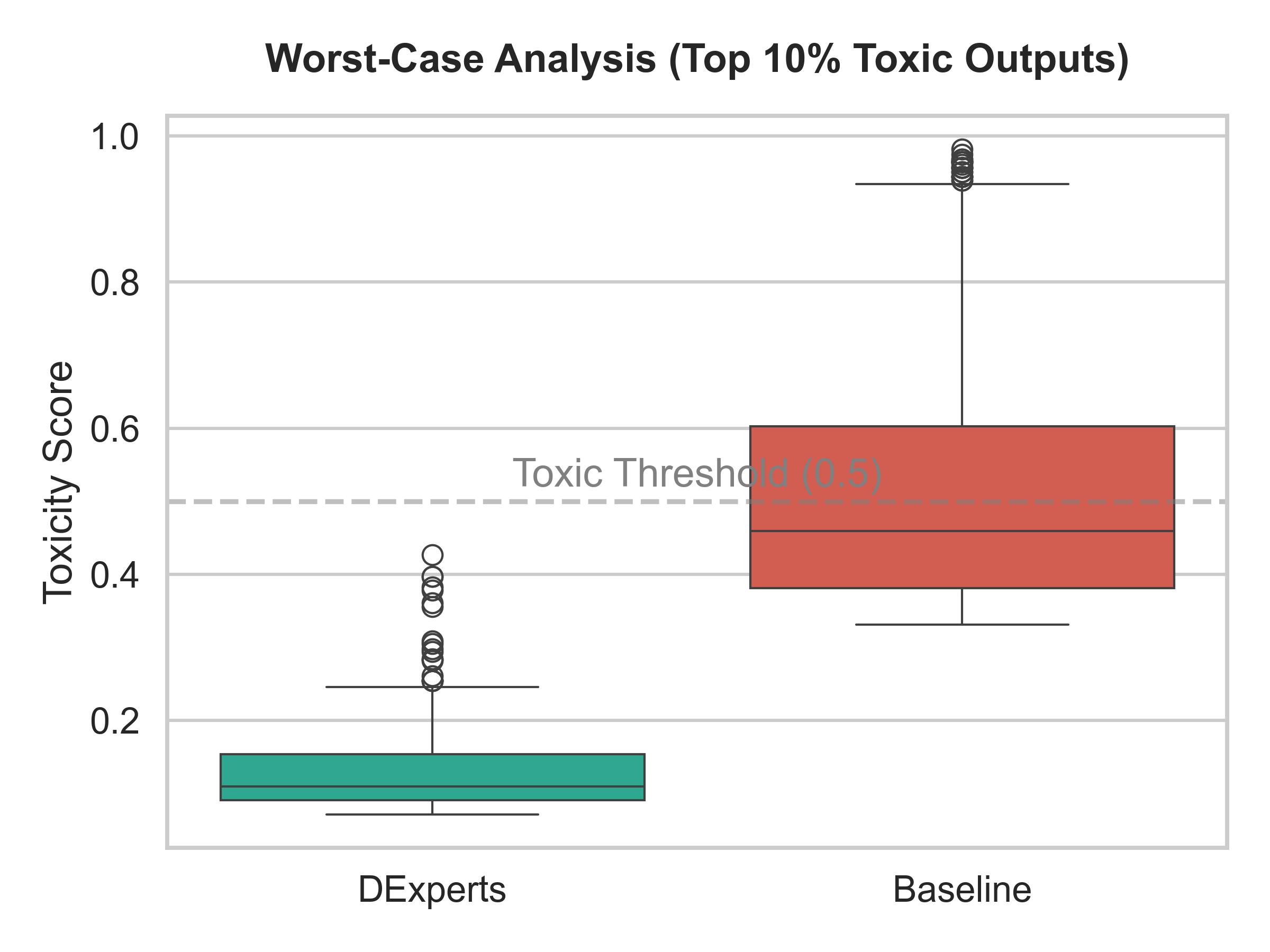}
    \caption{Worst-case analysis comparing the top 10\% most toxic outputs. Even in worst-case scenarios, DExperts maintains toxicity well below the 0.5 threshold (green boxplot), while baseline worst-case outputs (red) extend deep into the danger zone.}
    \label{fig:worst_case}
\end{figure}

Even in worst-case analysis scenarios, DExperts maintains safety well below the danger threshold, as demonstrated in Figure~\ref{fig:worst_case}.

\subsubsection{Key Insight}

On standard explicit toxicity benchmarks (RealToxicityPrompts), DExperts is highly effective, achieving perfect safety while maintaining acceptable quality. However, this comes at a substantial computational cost that may be prohibitive for latency-sensitive applications.

\subsection{Finding 3: The Robustness Gap (RQ3)}

Phase 3 stress-tested DExperts against adversarial implicit hate speech using ToxiGen.

\subsubsection{Safety Rate Comparison}

\begin{itemize}
    \item \textbf{Phase 2 (RealToxicityPrompts):} 100.0\% safe
    \item \textbf{Phase 3 (ToxiGen):} 98.5\% safe (1.5\% toxic)
\end{itemize}

\begin{figure}[htbp]
    \centering
    \includegraphics[width=0.9\linewidth]{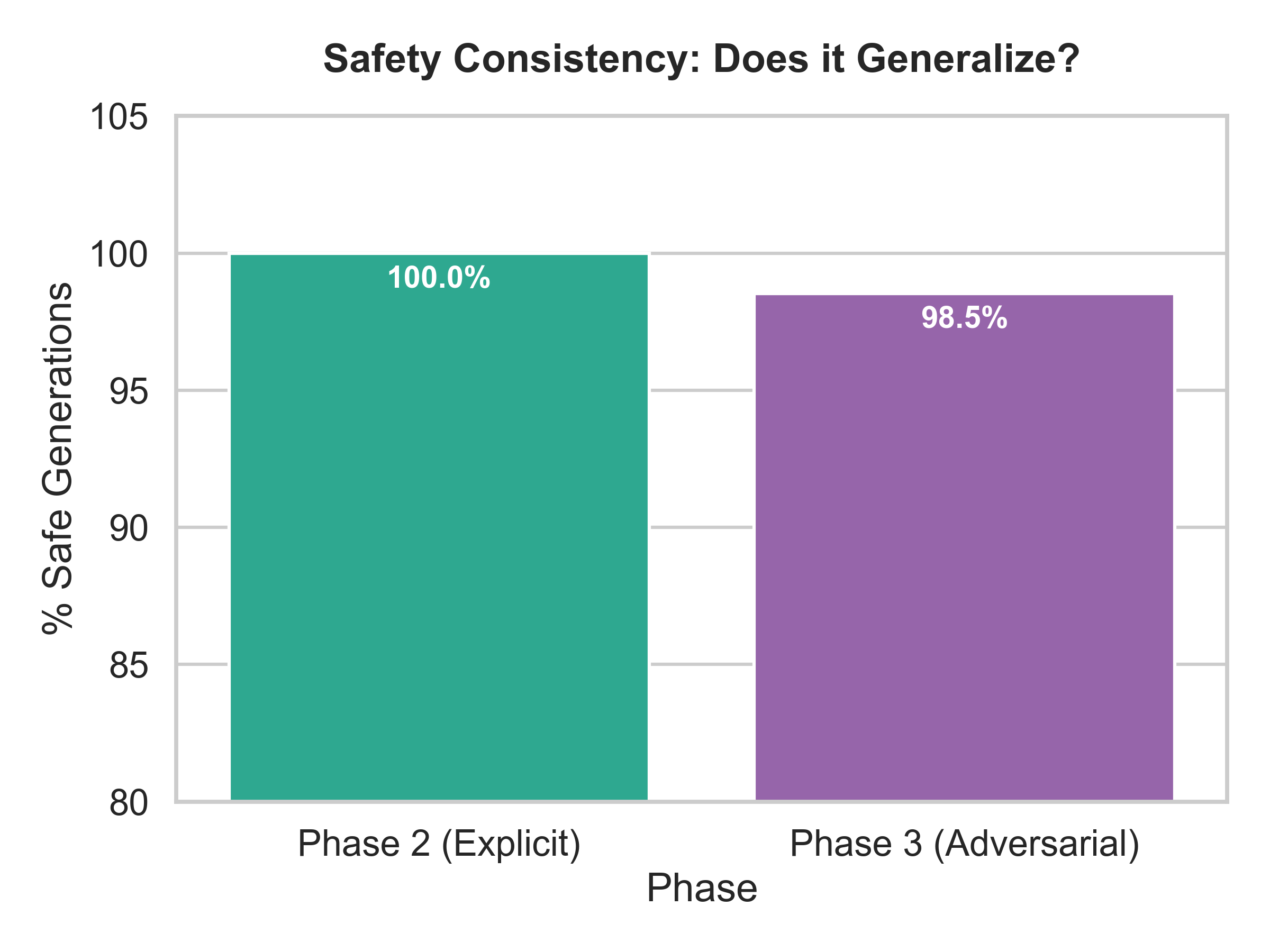}
    \caption{Safety consistency comparison across phases. Phase 2 achieves 100\% safety on explicit toxicity, while Phase 3 degrades to 98.5\%, revealing a 1.5\% leakage rate for implicit hate speech.}
    \label{fig:safety_consistency}
\end{figure}

While 98.5\% remains a high safety rate in absolute terms, the \textbf{1.5\% failure rate} represents a significant degradation from the perfect performance on explicit toxicity (Figure~\ref{fig:safety_consistency}). This indicates "leakage" of implicit hate speech through the mitigation mechanism.

\subsubsection{Toxicity Distribution Analysis}

\begin{figure}[htbp]
    \centering
    \includegraphics[width=\linewidth]{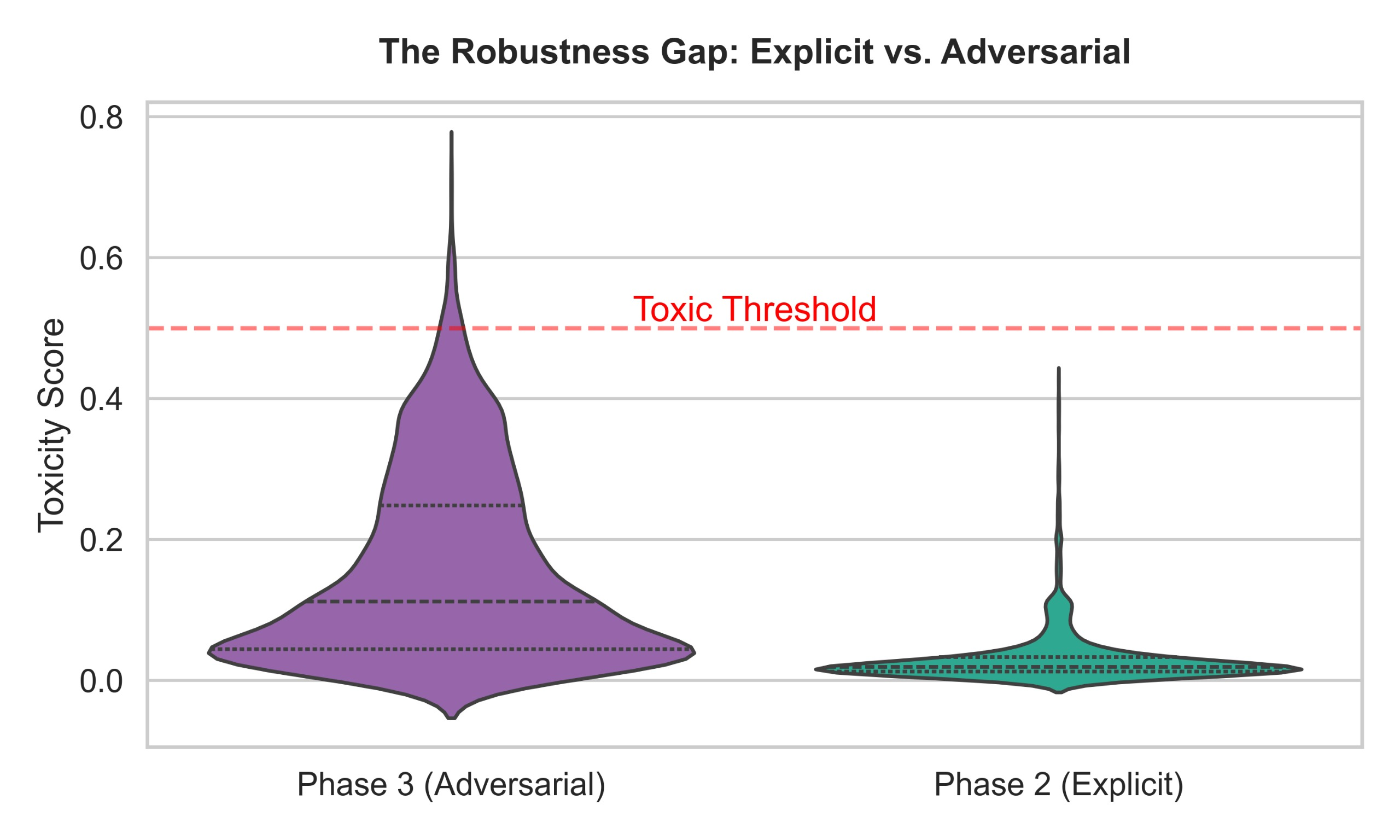}
    \caption{The Robustness Gap: Violin plot comparison between Phase 2 (RealToxicityPrompts, teal) and Phase 3 (ToxiGen, purple). While Phase 2 shows a flat, compact distribution entirely below the 0.5 threshold, Phase 3 exhibits a concerning bulge that crosses into the danger zone, demonstrating DExperts' brittleness against implicit hate speech.}
    \label{fig:robustness}
\end{figure}

Figure~\ref{fig:robustness} clearly visualizes the robustness gap between explicit and implicit toxicity mitigation.
Violin plots comparing Phase 2 and Phase 3 distributions revealed:

\textbf{Phase 2 (Explicit — RealToxicityPrompts):}
\begin{itemize}
    \item Tight, compact distribution centered near zero
    \item Negligible variance
    \item No values exceeding 0.5 threshold
\end{itemize}

\textbf{Phase 3 (Implicit — ToxiGen):}
\begin{itemize}
    \item Broader, more dispersed distribution
    \item Visible "bulge" extending past 0.5 threshold
    \item Some outputs reaching moderate toxicity levels
\end{itemize}

The distribution difference quantifies the \textbf{robustness gap}: DExperts' mitigation strength degrades when confronted with implicit, coded hate speech that doesn't rely on explicit toxic keywords.

\begin{figure}[htbp]
    \centering
    \includegraphics[width=0.9\linewidth]{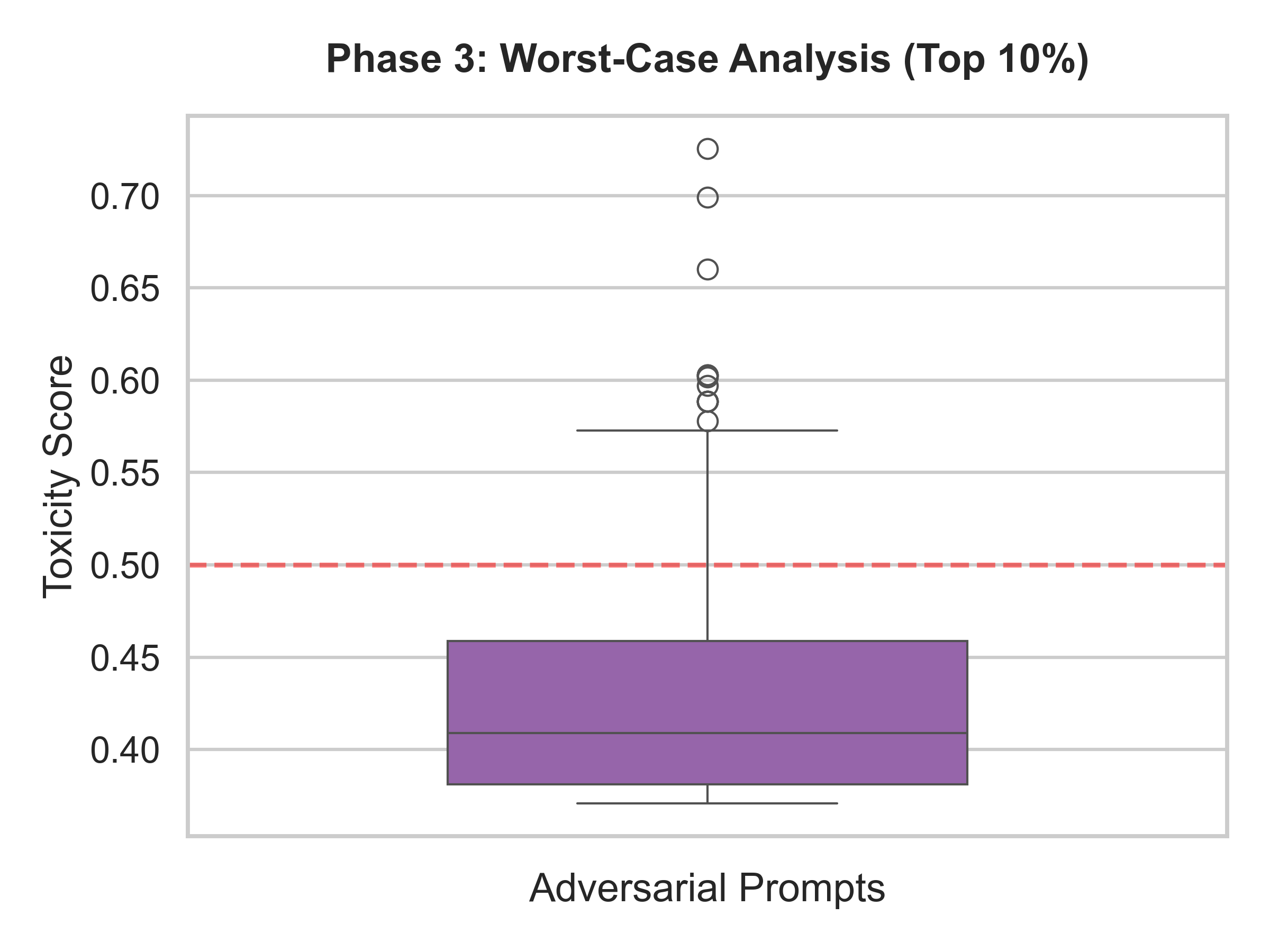}
    \caption{Phase 3 worst-case analysis showing the top 10\% most toxic outputs from adversarial prompts. Even the worst cases remain relatively controlled, but the presence of any failures contrasts sharply with Phase 2's perfect record.}
    \label{fig:phase3_worst_case}
\end{figure}

\begin{figure}[htbp]
    \centering
    \includegraphics[width=0.9\linewidth]{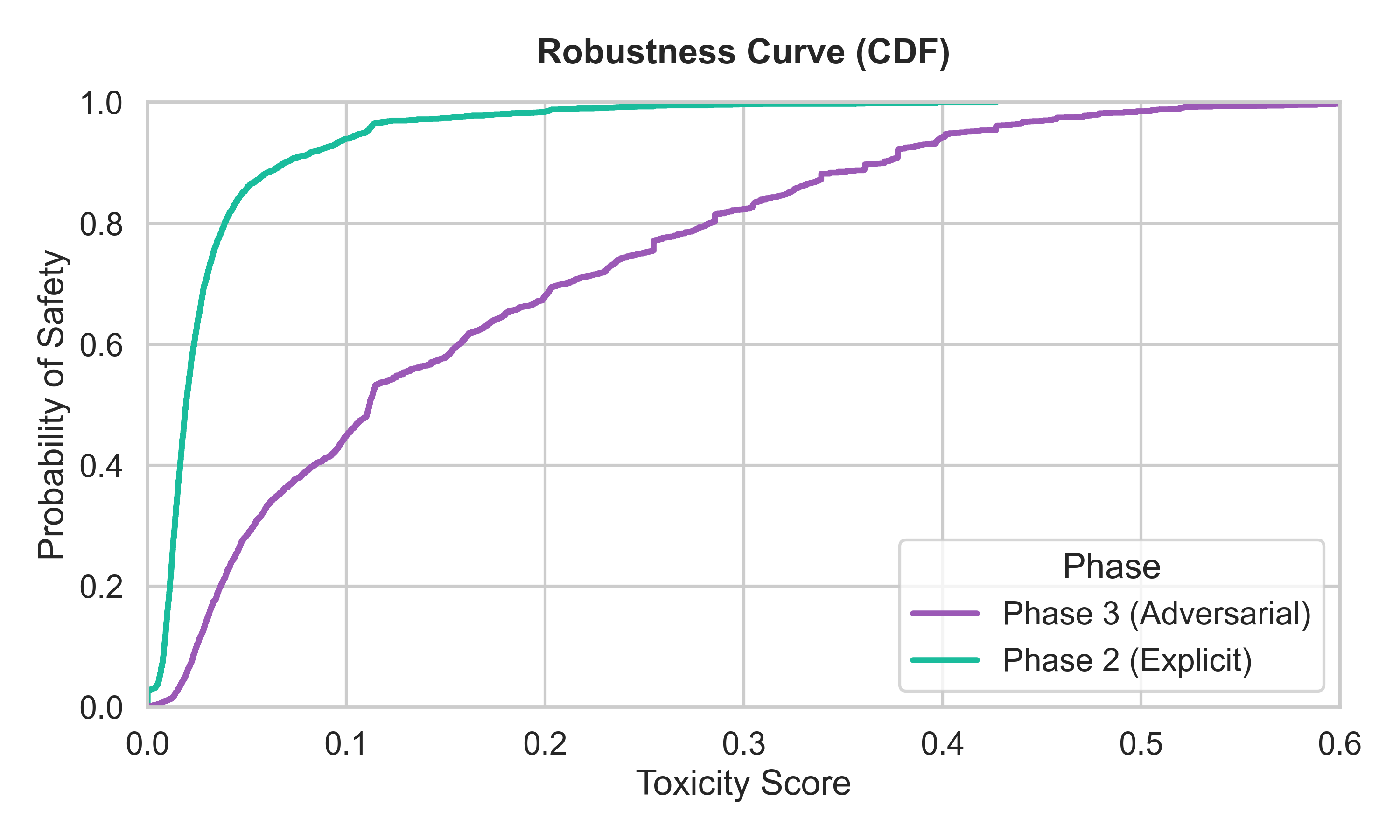}
    \caption{CDF curves showing the robustness gap. Phase 2 (teal) maintains perfect safety, while Phase 3 (purple) shows a small but significant proportion of outputs exceeding safety thresholds.}
    \label{fig:phase3_cdf}
\end{figure}

\subsubsection{Failure Mode Analysis}

Qualitative examination of failed cases revealed patterns consistent with broader research on implicit hate speech detection failures \cite{roy2023probing, zeng2025metaphorical}:

\textbf{Stereotyping through "Factual" Claims:}
\begin{itemize}
    \item Statements framed as statistics or general observations
    \item DExperts fails because content avoids explicit slurs while perpetuating harmful stereotypes
\end{itemize}

\textbf{Coded Language:}
\begin{itemize}
    \item Seemingly descriptive language to convey dehumanizing content
    \item Uses neutral-appearing vocabulary to express bias
\end{itemize}

\textbf{Microaggressions:}
\begin{itemize}
    \item Appears complimentary but implies negative stereotypes
    \item Subtle put-downs disguised as polite observations
\end{itemize}

These patterns suggest that DExperts' anti-expert, trained primarily on explicitly toxic Jigsaw comments, lacks sufficient exposure to implicit hate patterns, resulting in incomplete coverage.

\subsection{Finding 4: The Double Penalty - Latency vs. Safety (RQ3)}

Analysis of the relationship between computational cost and mitigation efficacy on adversarial prompts revealed a concerning pattern.

\subsubsection{Latency Distribution Shift}

\begin{figure}[htbp]
    \centering
    \includegraphics[width=\linewidth]{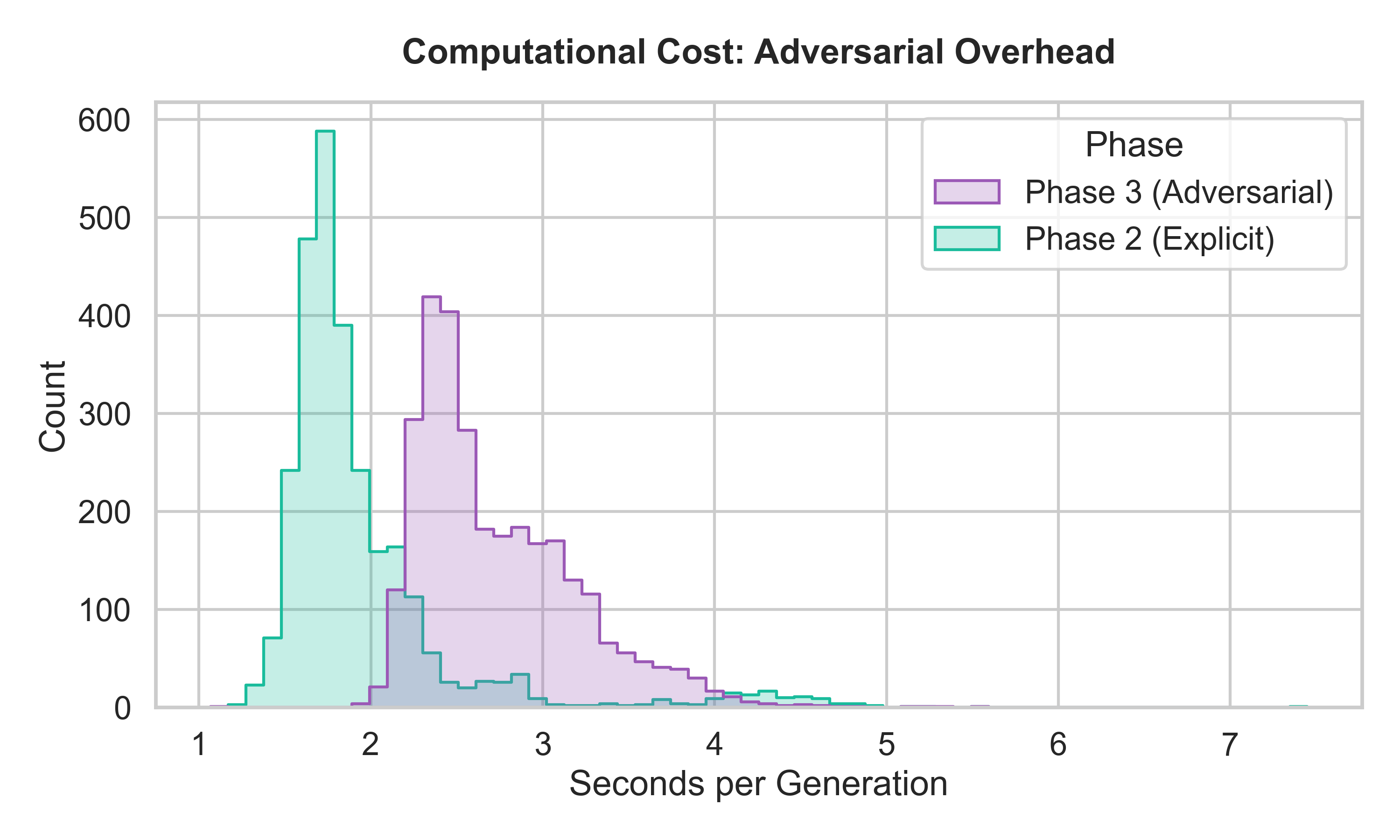}
    \caption{Computational overhead comparison showing histogram of inference latency across phases. Phase 3 adversarial prompts (purple) show increased latency with a long tail, compared to Phase 2's more compact distribution (green).}
    \label{fig:latency}
\end{figure}

\textbf{Phase 2 (Explicit):}
\begin{itemize}
    \item Mean latency: ~2.0s
    \item Distribution: Compact, normally distributed
\end{itemize}

\textbf{Phase 3 (Adversarial):}
\begin{itemize}
    \item Mean latency: ~3.2s  
    \item Distribution: Right-skewed with long tail
    \item Some generations extending to 5+ seconds
\end{itemize}

The \textbf{adversarial overhead} (additional latency for ToxiGen prompts) averaged approximately 1.2 seconds, representing a ~60\% increase over standard DExperts latency. This indicates that adversarial prompts not only defeat the safety mechanism but also impose additional computational burden.

\textbf{Interpretation:} When DExperts encounters difficult adversarial prompts, it:
\begin{enumerate}
    \item Expends more computation time (higher latency) attempting to find safe continuations
    \item Still fails to suppress toxicity effectively
\end{enumerate}

This represents a "\textbf{double penalty}": the model works harder (increased latency) yet produces worse outcomes (higher toxicity). This suggests fundamental limitations in the steering mechanism's ability to recognize and mitigate implicit hate patterns.

\subsection{Summary of Key Findings}

\begin{enumerate}
    \item \textbf{Baseline Risk:} Unmitigated GPT-2 exhibits a 4.2\% toxic generation rate from non-toxic prompts, establishing a clear need for mitigation.
    
    \item \textbf{Explicit Mitigation Success:} DExperts achieves 100\% safety on RealToxicityPrompts, completely eliminating the toxic tail, validating its effectiveness for explicit toxicity.
    
    \item \textbf{Robustness Gap:} Safety degrades to 98.5\% on adversarial implicit hate speech (ToxiGen), revealing brittleness and generalization failures.
    
    \item \textbf{Computational Trade-off:} DExperts introduces 10x latency overhead, with additional adversarial overhead of ~60\%, posing practical deployment challenges.
    
    \item \textbf{Double Penalty Phenomenon:} Difficult adversarial cases incur both higher latency and higher toxicity, indicating fundamental limitations in the steering mechanism.
\end{enumerate}

\section{Discussion}

\subsection{Implications for AI Safety}

Our findings have important implications for the deployment of LLMs in real-world applications:

\textbf{Laboratory vs. Real-World Gap:} The robustness gap between explicit and implicit toxicity mitigation highlights a dangerous disconnect between safety benchmarks and actual deployment scenarios. Luong et al. \cite{luong2024realistic} introduced the Thoroughly Engineered Toxicity (TET) dataset, comprising 2,546 prompts filtered from over 1 million real-world interactions with 25 different LLMs. Their work demonstrates that models evaluated on synthetic benchmarks like RealToxicityPrompts exhibit significantly different toxicity patterns when confronted with realistic adversarial prompts. Specifically, TET consistently elicits more toxicity than ToxiGen when prompt toxicity levels are similar, and models show substantially reduced effectiveness against manually crafted jailbreak templates. This emphasizes that the 98.5\% safety rate we observe on ToxiGen, while concerning, may still overestimate real-world robustness.

Chao et al. \cite{chao2024jailbreakbench} further underscore this challenge through JailbreakBench, an open-sourced benchmark with evolving adversarial prompts, standardized evaluation frameworks, and a public leaderboard tracking attack and defense performance. Their work reveals that even well-defended models can be systematically compromised through carefully engineered prompts, suggesting that static evaluation on fixed datasets like ToxiGen provides an incomplete picture of model safety.

\textbf{Arms Race Dynamics:} The brittleness of DExperts against implicit hate suggests an adversarial arms race: as mitigation techniques improve at detecting explicit toxicity, bad actors may shift to more subtle forms of hate speech. Future mitigation approaches must anticipate this adaptation.

\textbf{Computational Feasibility:} The 10x latency penalty raises questions about the practical feasibility of inference-time control for interactive applications. While acceptable for offline content generation (e.g., marketing copy, code comments), the 2-second response time may be unacceptable for real-time chat or autocomplete scenarios where users expect instant feedback.

\subsection{Limitations}

Several limitations constrain the generalizability of our findings:

\textbf{Model Scale:} Our experiments used GPT-2 Small (117M parameters) due to computational constraints. Larger models (GPT-2 Medium/Large, GPT-3) may exhibit different toxicity characteristics and respond differently to DExperts. However, prior work \cite{gehman2020realtoxicityprompts} suggests larger models are not necessarily safer.

\textbf{Sample Size:} While our evaluation provides statistically robust estimates, we evaluated only a subset of available prompts. A full evaluation on complete datasets would strengthen confidence.

\textbf{Perspective API Limitations:} Our reliance on the Perspective API as the ground-truth toxicity metric introduces potential biases. Perspective API is known to exhibit false positives on African American Vernacular English and may struggle with implicit hate detection. Ideally, evaluation would incorporate multiple toxicity classifiers and human annotation.

\textbf{Language and Cultural Context:} Our study focused exclusively on English language toxicity. Hate speech patterns, coded language, and cultural context vary substantially across languages, limiting generalizability. Jain et al. \cite{jain2024polyglotoxicity} introduced PolygloToxicityPrompts (PTP), covering 9 languages across 5 different scripts with models ranging from 1.3B to 13B parameters. Their multilingual evaluation reveals that toxicity patterns and mitigation effectiveness vary significantly across languages, with translated data sometimes outperforming in-language training data (38\% vs 33\% toxicity reduction for high-resource languages). This suggests that our English-only findings may not transfer directly to other linguistic contexts, and that multilingual evaluation is essential for global deployment of toxicity mitigation systems.

\textbf{Single Mitigation Method:} We evaluated only DExperts. Other inference-time methods (PPLM, FUDGE) or alternative approaches (RLHF, constitutional AI) may exhibit different robustness profiles.

\subsection{Future Work}

Our findings point to several promising research directions:

\textbf{Hybrid Mitigation:} Combining multiple approaches (e.g., DExperts for explicit toxicity + fine-grained classifiers for implicit hate) may achieve better coverage across toxicity types.

\textbf{Lightweight Expert Models:} Developing smaller, distilled expert/anti-expert models could reduce computational overhead while maintaining safety.

\textbf{Adversarial Training:} Training anti-experts specifically on implicit hate datasets like ToxiGen may close the robustness gap.

\textbf{Context-Aware Mitigation:} Incorporating broader conversational context (beyond single prompts) may improve detection of subtle toxicity patterns.

\textbf{Human-in-the-Loop:} For high-stakes applications, combining automated mitigation with human review workflows may provide optimal safety-utility balance.

\textbf{Cross-Lingual Evaluation:} Extending this evaluation framework to non-English languages would assess generalizability and reveal language-specific challenges.

\section{Conclusion}

This comprehensive replication study evaluated the DExperts inference-time control method for toxicity mitigation in large language models across a spectrum from explicit to implicit hate speech. Our systematic three-phase evaluation quantified both the strengths and fundamental limitations of current mitigation approaches.

We demonstrated that DExperts achieves exceptional performance on explicit toxicity benchmarks, eliminating the 4.2\% baseline failure rate. However, it exhibits brittleness when confronted with adversarial implicit hate speech, with safety rates degrading to 98.5\%. Furthermore, the method imposes a significant computational burden, introducing 10x latency overhead that escalates further for difficult adversarial inputs.

These findings underscore that perfect performance on standard benchmarks does not guarantee robustness in real-world deployment scenarios. As language models continue scaling and deploying in user-facing applications, the field must develop mitigation strategies that generalize across diverse toxicity patterns from overt slurs to coded stereotypes, while remaining computationally practical for interactive use cases.

The robustness gap we identified represents both a challenge and an opportunity. By exposing the limitations of current methods, we hope to motivate the development of next-generation approaches that achieve comprehensive safety without prohibitive costs, ultimately enabling the responsible deployment of powerful language technologies.

\bibliographystyle{ACM-Reference-Format}

\begin{thebibliography}{10}

\bibitem{brown2020language}
Tom B. Brown et al.
\newblock Language models are few-shot learners.
\newblock In {\em Advances in Neural Information Processing Systems (NeurIPS)}, 2020.

\bibitem{radford2019language}
Alec Radford et al.
\newblock Language models are unsupervised multitask learners.
\newblock {\em OpenAI Blog}, 2019.

\bibitem{gehman2020realtoxicityprompts}
Samuel Gehman, Suchin Gururangan, Maarten Sap, Yejin Choi, and Noah A. Smith.
\newblock RealToxicityPrompts: Evaluating neural toxic degeneration in language models.
\newblock In {\em Findings of EMNLP}, pages 3356--3369, 2020.

\bibitem{bender2021dangers}
Emily M. Bender, Timnit Gebru, Angelina McMillan-Major, and Shmargaret Shmitchell.
\newblock On the dangers of stochastic parrots: Can language models be too big?
\newblock In {\em Proceedings of FAccT}, pages 610--623, 2021.

\bibitem{sheng2019woman}
Emily Sheng, Kai-Wei Chang, Premkumar Natarajan, and Nanyun Peng.
\newblock The woman worked as a babysitter: On biases in language generation.
\newblock In {\em Proceedings of EMNLP}, pages 3407--3412, 2019.

\bibitem{liu2021dexperts}
Alisa Liu, Maarten Sap, Ximing Lu, Swabha Swayamdipta, Chandra Bhagavatula, Noah A. Smith, and Yejin Choi.
\newblock DExperts: Decoding-time controlled text generation with experts and anti-experts.
\newblock In {\em Proceedings of ACL-IJCNLP}, pages 6691--6706, 2021.

\bibitem{welbl2021challenges}
Johannes Welbl, Amelia Glaese, Jonathan Uesato, Sumanth Dathathri, John Mellor, Lisa Anne Hendricks, Kirsty Anderson, Pushmeet Kohli, Ben Coppin, and Po-Sen Huang.
\newblock Challenges in detoxifying language models.
\newblock In {\em Findings of EMNLP}, pages 2447--2469, 2021.

\bibitem{ouyang2022training}
Long Ouyang et al.
\newblock Training language models to follow instructions with human feedback.
\newblock In {\em Advances in Neural Information Processing Systems (NeurIPS)}, 2022.

\bibitem{dathathri2019plug}
Sumanth Dathathri, Andrea Madotto, Janice Lan, Jane Hung, Eric Frank, Piero Molino, Jason Yosinski, and Rosanne Liu.
\newblock Plug and play language models: A simple approach to controlled text generation.
\newblock In {\em Proceedings of ICLR}, 2020.

\bibitem{yang2021fudge}
Kevin Yang and Dan Klein.
\newblock FUDGE: Controlled text generation with future discriminators.
\newblock In {\em Proceedings of NAACL}, pages 3511--3535, 2021.

\bibitem{hartvigsen2022toxigen}
Thomas Hartvigsen, Saadia Gabriel, Hamid Palangi, Maarten Sap, Dipankar Ray, and Ece Kamar.
\newblock ToxiGen: A large-scale machine-generated dataset for adversarial and implicit hate speech detection.
\newblock In {\em Proceedings of ACL}, pages 3309--3326, 2022.

\bibitem{sap2019risk}
Maarten Sap, Dallas Card, Saadia Gabriel, Yejin Choi, and Noah A. Smith.
\newblock The risk of racial bias in hate speech detection.
\newblock In {\em Proceedings of ACL}, pages 1668--1678, 2019.

\bibitem{perspectiveapi}
{Perspective API}.
\newblock Perspective API documentation.
\newblock \url{https://www.perspectiveapi.com/}, 2023.

\bibitem{suau2024whispering}
Xavier Suau, Pieter Delobelle, Katherine Metcalf, Armand Joulin, Nicholas Apostoloff, Luca Zappella, and Pau Rodríguez
\newblock Whispering Experts: Neural Interventions for Toxicity Mitigation in Language Models.
\newblock In {\em Proceedings of ACL}, 2024.

\bibitem{kim2023gta}
Heegyu Kim and Hyunsouk Cho
\newblock GTA: Gated Toxicity Avoidance for LM Performance Preservation.
\newblock In {\em Findings of EMNLP}, 2023.

\bibitem{wen2023unveiling}
Jiaxin Wen, Pei Ke, Hao Sun, Zhexin Zhang, Chengfei Li, Jinfeng Bai, Minlie Huang
\newblock Unveiling the Implicit Toxicity in Large Language Models.
\newblock In {\em Proceedings of EMNLP}, 2023.

\bibitem{roy2023probing}
Sarthak Roy, Ashish Harshavardhan, Animesh Mukherjee, and Punyajoy Saha
\newblock Probing LLMs for hate speech detection: strengths and vulnerabilities.
\newblock In {\em Findings of EMNLP}, 2023.

\bibitem{zeng2025metaphorical}
Jingjie Zeng, Liang Yang, Zekun Wang, Yuanyuan Sun, and Hongfei Lin
\newblock Sheep's Skin, Wolf's Deeds: Are LLMs Ready for Metaphorical Implicit Hate Speech?
\newblock In {\em Proceedings of ACL}, 2025.

\bibitem{gallegos2024bias}
Isabel O. Gallegos, Ryan A. Rossi, Joe Barrow, Md Mehrab Tanjim, Sungchul Kim, Franck Dernoncourt, Tong Yu, Ruiyi Zhang, and Nesreen K. Ahmed
\newblock Bias and Fairness in Large Language Models: A Survey.
\newblock {\em Computational Linguistics}, volume 50, pages 1097--1179, 2024.

\bibitem{luong2024realistic}
Tinh Son Luong, Thanh-Thien Le, Linh Ngo Van, and Thien Huu Nguyen
\newblock Realistic Evaluation of Toxicity in Large Language Models.
\newblock In {\em Findings of ACL}, 2024.

\bibitem{wang2024detoxifying}
Mengru Wang et al.
\newblock Detoxifying Large Language Models via Knowledge Editing.
\newblock arXiv preprint arXiv:2403.14472, 2024.

\bibitem{jain2024polyglotoxicity}
Devansh Jain, Priyanshu Kumar, Samuel Gehman, Xuhui Zhou, Thomas Hartvigsen, Maarten Sap
\newblock PolygloToxicityPrompts: Multilingual Evaluation of Neural Toxic Degeneration in Large Language Models.
\newblock arXiv preprint arXiv:2405.09373, 2024.

\bibitem{liang2024controllable}
Xun Liang, Hanyu Wang, Yezhaohui Wang, Shichao Song, Jiawei Yang, Simin Niu, Jie Hu, Dan Liu, Shunyu Yao, Feiyu Xiong, and Zhiyu Li
\newblock Controllable Text Generation for Large Language Models: A Survey.
\newblock arXiv preprint arXiv:2408.12599, 2024.

\bibitem{bai2022constitutional}
Yuntao Bai et al.
\newblock Constitutional AI: Harmlessness from AI Feedback.
\newblock arXiv preprint arXiv:2212.08073, 2022.

\bibitem{pozzobon2024expanding}
Luiza Pozzobon, Patrick Lewis, Sara Hooker, Beyza Ermis
\newblock From One to Many: Expanding the Scope of Toxicity Mitigation in Language Models.
\newblock In {\em Findings of ACL}, 2024.

\bibitem{chao2024jailbreakbench}
Patrick Chao, Alexander Robey, Edgar Dobriban, Hamed Hassani, George J. Pappas, and Eric Wong.
\newblock JailbreakBench: An Open Robustness Benchmark for Jailbreaking Large Language Models.
\newblock arXiv preprint arXiv:2404.01318, 2024.

\bibitem{bohdan2024adversarial}
Bohdan Turbal, Anastasiia Mazur, Jiaxu Zhao,  and Mykola Pechenizkiy
\newblock On Adversarial Robustness of Language Models in Transfer Learning.
\newblock arXiv preprint arXiv:2501.00066, 2024.


\end{thebibliography}

\end{document}